\documentclass{article}

\usepackage[OT1]{fontenc}
\usepackage{iclr2027_conference,times}
\usepackage{amsmath,amsfonts,bm}

\def\eqref#1{equation~\ref{#1}}
\def\1{\bm{1}}

\DeclareMathAlphabet{\mathsfit}{\encodingdefault}{\sfdefault}{m}{sl}
\SetMathAlphabet{\mathsfit}{bold}{\encodingdefault}{\sfdefault}{bx}{n}

\usepackage{amsmath,amssymb,bm}
\usepackage{booktabs}
\usepackage{graphicx}
\usepackage{xcolor}
\usepackage{placeins}
\usepackage{hyperref}
\usepackage{url}
\usepackage{multirow}

\newcounter{algorithm}
\renewcommand{\thealgorithm}{\arabic{algorithm}}
\newcommand{\algorithmcaption}[1]{%
  \hrule\vspace{0.3em}%
  \noindent\textbf{Algorithm \thealgorithm:} #1\par\vspace{0.3em}\hrule}

\newcommand{\method}{EbbFlow}

\newcommand{\hmax}{\ensuremath{h_{\max}}}

\title{EbbFlow: PINN-Inspired Progressive\\
\mbox{Curriculum} for MeanFlow Training}

\author{%
  {\normalsize\itshape Mengfan Yuan\textsuperscript{1,2}, Xunheng Kuang\textsuperscript{3},
  Jianbin Chen\textsuperscript{1}, Lei He\textsuperscript{4}}\\[5pt]
  {\small\textsuperscript{1} Beijing Institute of Technology (bit.edu.cn)}\\[2pt]
  {\small\textsuperscript{2} Tsinghua University (mail.tsinghua.edu.cn)}\\[2pt]
  {\small\textsuperscript{3} Northwestern Polytechnical University, Xi'an, China}\\[2pt]
  {\small\textsuperscript{4} Tsinghua University (tsinghua.edu.cn)}%
}

\makeatletter
\renewcommand{\@maketitle}{\vbox{\hsize\textwidth
  {\LARGE\sc\@title\par}
  \vskip 12pt
  \noindent\begin{tabular}[t]{@{}l@{}}\@author\end{tabular}
  \vskip 0.3in minus 0.1in}}
\makeatother
\ClearShipoutPicture
\renewcommand{\headrulewidth}{0pt}
\hypersetup{%
  pdftitle={EbbFlow: PINN-Inspired Progressive Curriculum for MeanFlow Training},
  pdfauthor={Mengfan Yuan, Xunheng Kuang, Jianbin Chen, Lei He}%
}

\begin{document}
\maketitle

\begin{abstract}
MeanFlow learns interval-average velocity for few-step generation, but its
fixed temporal sampler mixes diagonal and interior supervision throughout
training. We interpret temporal coordinates as collocation points: diagonal
samples provide a stochastic boundary target, whereas finite intervals supply
a differential-consistency target. The diagonal fraction controls how often
the consistency component appears, while interval length enters its gradient
through an explicit factor. Motivated by this distinction, EbbFlow uses
Temporal Supervision Allocation (TSA) to schedule the diagonal fraction and
interval geometry. It starts with diagonal supervision, then admits interior
samples and expands their maximum interval; a left-anchored interval law
preserves coverage across time. The per-sample loss, target, backbone, and
inference procedure remain unchanged. In matched class-conditional ImageNet
$256\times256$ runs without classifier-free guidance (DiT-B/2, one training
seed per method, 400k updates), EbbFlow achieves 50k-sample FID of 39.65 at
one network evaluation and 36.02 at two-evaluation ODE sampling, versus 44.82
and 39.86 for MeanFlow. Gradient probes show a mid-training difference in
component geometry. Ablations support allocation scheduling and left-anchored
intervals, although a fixed full-horizon variant performs better at one
evaluation. These results suggest that temporal sampling is a useful training
control whose benefits depend on the inference protocol.
\end{abstract}

\section{Introduction}
\label{sec:introduction}

Generative flow models transport noise to data by learning a velocity field.
MeanFlow \citep{geng2025mean} replaces instantaneous velocity with an average
velocity over a finite interval, making one- and two-step generation possible
without changing the backbone. Its training coordinates nevertheless mix two
qualitatively different cases: diagonal points $(r,t)=(t,t)$ recover ordinary
flow-matching supervision, whereas interior points $r<t$ impose a
self-referential finite-interval identity. The default ImageNet recipe fixes
the diagonal fraction at $0.75$ and exposes the model to the full interval
range from the first update. These constants do not adapt as the model and its
derivative-based target evolve.

We expose this choice by viewing temporal coordinates as collocation points on
the triangle $0\leq r\leq t\leq1$. This reveals two independent controls. The
diagonal fraction $\rho$ determines \emph{how many} samples receive only the
stable boundary condition; the interval length $h=t-r$ determines \emph{how
strongly} each interior sample carries the differential component, through an
explicit factor of $h$. Interval-binned gradient probes confirm that the two
components are antagonistic and that their directional conflict and scale
imbalance vary with both interval length and training progress. A fixed
collocation measure therefore presents the same mixture of constraints even
when their optimization geometry is changing.

Motivated by curriculum remedies for ill-conditioned physics-informed neural
networks \citep[PINNs;][]{krishnapriyan2021pinn}, we propose \method{}, which uses
Temporal Supervision Allocation (TSA) to schedule its training coordinates. TSA first emphasizes the
diagonal and short intervals, then progressively admits more interior samples
and expands the interval horizon. A left-anchored construction preserves
global coverage of time throughout the curriculum. Only the sampling measure
changes: the MeanFlow per-sample loss, target construction, backbone, guidance, and
inference procedure remain untouched. This also distinguishes TSA from
AlphaFlow \citep{zhang2025alphaflow}, which changes the target construction;
the two mechanisms are conceptually orthogonal.
Short-to-long interval curricula also appear in gap-dependent weighting
\citep{kim2026understanding} and pretrained speech enhancement
\citep{wang2025meanse}. Our contribution is to separate and jointly schedule
diagonal allocation and interval geometry within the sampling measure.

Our contributions are threefold. First, we formulate MeanFlow training as a
boundary--interior collocation problem and separate allocation from interval
geometry. Second, we connect the interval length to consistency-gradient scale
and measure the resulting conflict throughout training. Third, in a matched
single-seed ImageNet $256\times256$ comparison with DiT-B/2, EbbFlow improves FID
from $44.82$ to $39.65$ at 1-NFE and from $39.86$ to $36.02$ at 2-NFE.
Ablations show that both scheduling axes matter, while a fixed full-horizon
left-anchored sampler remains better at 1-NFE. We accordingly present EbbFlow as a
complementary optimization recipe, not a universally optimal schedule.

\begin{figure}[t]
  \centering
  \includegraphics[width=0.98\linewidth]{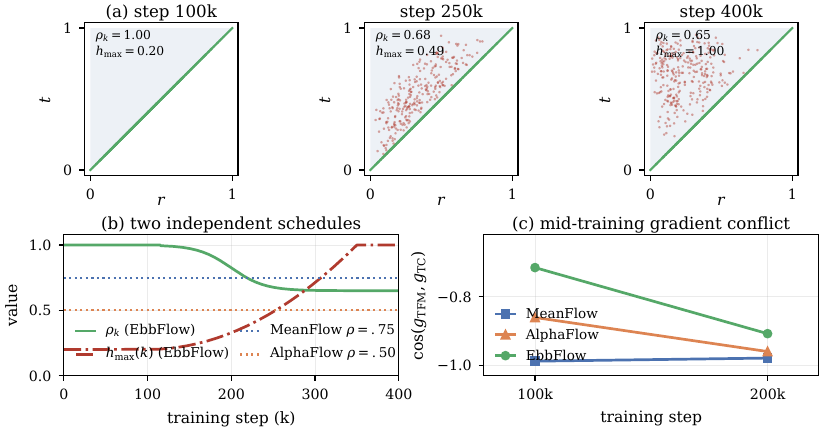}
  \caption{EbbFlow overview: (a) sampled temporal coordinates;
  (b) diagonal allocation $\rho_k$ and horizon $h_{\max}(k)$;
  (c) mean component-gradient cosine at 100k and 200k.
  Figure~\ref{fig:three-way-gradient-path} gives the full 100k--400k comparison.}
  \label{fig:teaser-collocation}
\end{figure}

\section{Related Work}
\label{sec:related}

\paragraph{Few-step generation and flow maps.}
Diffusion and flow-matching models learn transport from noise to data
\citep{ho2020ddpm,song2021sde,lipman2022flow,liu2023rectified}.
Latent representations and transformer backbones make these models practical
for image synthesis \citep{rombach2022high,peebles2023scalable,esser2024scaling}.
Few-step generation reduces repeated network evaluations through teacher
distillation \citep{salimans2022progressive,yin2024one}, trajectory consistency
\citep{song2023consistency,song2024improved,geng2025consistency,lu2025scm},
or learned two-time flow maps
\citep{kim2024consistency,boffi2025flow,frans2025shortcut,zhou2025imm}.
EbbFlow instead retains MeanFlow's transport identity and changes its
training coordinates.

\paragraph{MeanFlow and its variants.}
MeanFlow parameterizes transport by an average velocity whose differential
identity reduces to flow matching at $r=t$ \citep{geng2025mean}.
SplitMeanFlow replaces differentiation with interval-splitting consistency
\citep{guo2025splitmeanflow}; improved MeanFlow changes the loss
parameterization and guidance conditioning \citep{geng2025imf}.
Other variants modulate target gradients \citep{you2025modular}, decouple
timestep conditioning \citep{lee2025decoupled}, model ODE solutions
\citep{luo2026soflow}, straighten couplings \citep{zhang2025curvature},
or reduce target variance \citep{lu2026variance}. Discrete MeanFlow extends
generation to finite state spaces through transition kernels
\citep{khan2026discrete}. EbbFlow keeps the continuous MeanFlow target and
parameterization fixed, scheduling where its identity is enforced.

\paragraph{Training dynamics and MeanFlow curricula.}
AlphaFlow identifies opposed trajectory-flow-matching and consistency
gradients and anneals the target between these objectives
\citep{zhang2025alphaflow}. Temporal Equilibrium MeanFlow
\citep[TEMF;][]{tu2026temporal} addresses temporal scale imbalance through a fixed
interval-dependent loss weight and a sigmoid schedule that increases the
off-diagonal fraction, shifting from boundary stabilization to trajectory
learning. Thus, boundary-to-interior allocation scheduling already appears
in prior work. \citet{kim2026understanding} show that accurate
instantaneous velocities support average-velocity learning: short gaps help
their formation, whereas large gaps can disrupt it. Their method accelerates
instantaneous-velocity learning and progressively relaxes gap-dependent loss
weighting. Related curricula introduce trajectory-based mid-training
\citep{hu2025cmt}, anneal a target discretization step
\citep{hsu2026discrete}, or warm up gradient modulation
\citep{you2025modular}. MeanSE initializes from a pretrained speech flow model
and increases the maximum interval in stages from $0.2$ to $1$
\citep{wang2025meanse}. Short-to-long curricula are therefore also established.
EbbFlow studies allocation together with a left-anchored interval law and a
progressively expanded sampling horizon, retaining the baseline loss rather
than introducing TEMF's interval weight. Its collocation view and interval-binned
gradient diagnostics examine these sampling choices; the present experiments
do not establish superiority over TEMF.

\paragraph{Timestep sampling, loss weighting, and curricula.}
Diffusion training distributes effort over noise levels through loss weights
\citep{hang2023minsnr} or sampling densities, including the logit-normal law
used in rectified-flow training \citep{esser2024scaling}.
With matching density-ratio weights and sufficient proposal support,
importance resampling preserves the population objective but changes estimator
variance \citep{kingma2023understanding}. TSA instead changes the sampling
measure without compensation; Section~\ref{sec:design-choices} discusses
its relation to adaptive loss normalization. Related work treats diffusion
timesteps as interacting tasks
\citep{go2023negativetransfer}, schedules denoising difficulty
\citep{kim2025dtd,bengio2009curriculum}, and anneals consistency discretization
\citep{song2024improved,geng2025consistency}. TSA schedules both allocation
and interval geometry on MeanFlow's two-dimensional coordinate triangle.

\paragraph{PINN optimization and collocation.}
PINNs jointly fit boundary data and interior differential residuals
\citep{raissi2019pinn}. Their gradients can be imbalanced
\citep{wang2021gradientpathologies}, and differential constraints can produce
ill-conditioned loss landscapes \citep{krishnapriyan2021pinn,rathore2024challenges}.
Remedies include curriculum warm starts \citep{krishnapriyan2021pinn},
causality-respecting training \citep{wang2024causality,mattey2022novel}, and
adaptive collocation \citep{wu2023comprehensive}. Physics Informed Distillation
learns the solution of a teacher diffusion model's probability-flow ODE
\citep{tee2024pid}; EbbFlow instead applies the optimization analogy to
training MeanFlow from scratch. We examine
coordinate-dependent gradient interactions in
Figure~\ref{fig:gradient-conflict-by-interval}, using diagnostics also found in
multi-task learning \citep{chen2018gradnorm,yu2020pcgrad,fifty2021tag}.

\section{Analyzing the Temporal Sampling of MeanFlow}
\label{sec:background}\label{sec:analysis}

MeanFlow is never trained on a single coordinate. Every update draws a batch
of temporal pairs $(r,t)$, and the ImageNet recipe fixes how they are drawn:
$75\%$ of the batch sits on the diagonal $r=t$, while the remaining pairs
cover the full interval range $h=t-r\in(0,1)$ from the first step onward
\citep{geng2025mean}. \citet{zhang2025alphaflow} ask why so much of the budget
is spent on the diagonal. We ask a complementary question: \emph{why should
either constant be fixed at all?}

\subsection{Preliminaries}
\label{sec:prelim}

\paragraph{Flow matching.}
On the linear path $x_t=(1-t)\,x+t\,\epsilon$ with data $x$ and noise
$\epsilon$, Flow Matching \citep{lipman2022flow} regresses the conditional
velocity $\epsilon-x$ with a network $v_\theta(x_t,t)$, whose minimizer is the
marginal field $v(x_t,t)=\mathbb{E}[\epsilon-x\mid x_t]$. This target does not
depend on $\theta$; the MeanFlow target below does.

\paragraph{MeanFlow.}
MeanFlow \citep{geng2025mean} models the average velocity over a finite
interval $[r,t]$ with $t>r$ along a marginal-velocity characteristic,
\begin{equation}
u(x_t,r,t)\;\triangleq\;\frac{1}{t-r}\int_{r}^{t} v(x_\tau,\tau)\,d\tau ,
\label{eq:meanflow-avg}
\end{equation}
with $u(x_t,t,t)=v(x_t,t)$ by continuous extension. A network
$u_\theta(x_t,r,t)$ predicts this quantity directly. Direct integration is
intractable during training, so MeanFlow differentiates
Equation~\ref{eq:meanflow-avg} with respect to $t$ to obtain the MeanFlow
identity
\begin{equation}
u(x_t,r,t)=v(x_t,t)-(t-r)\,\frac{d}{dt}u(x_t,r,t),
\label{eq:meanflow-identity}
\end{equation}
where the total derivative is computed with a Jacobian--vector product (JVP).
One-step generation follows from $x_0=x_1-u_\theta(x_1,0,1)$, i.e.\ from the
single coordinate $(r,t)=(0,1)$.

Write $h=t-r$ for the interval length and $\widetilde v_t$ for the sampled
training velocity ($\epsilon-x$ without guidance, or the velocity returned by
the baseline's training-time guidance construction). Let
$\widetilde{\mathcal D}_t=\partial_t+\widetilde v_t^\top\nabla_{x_t}$ be the
derivative along the sampled characteristic, evaluated with a JVP. MeanFlow
trains $u_\theta$ against the stop-gradient target
\begin{equation}
y_\theta(x_t,r,t)=\operatorname{sg}\!\bigl[\,\widetilde v_t
 -h\,\widetilde{\mathcal D}_t u_\theta(x_t,r,t)\bigr],
\qquad
\ell_{\mathrm{MF}}=\bigl\lVert u_\theta(x_t,r,t)-y_\theta(x_t,r,t)\bigr\rVert_2^2 .
\label{eq:stopped-target}
\end{equation}
Two properties of Equation~\ref{eq:stopped-target} are used repeatedly
below. First, away from $r=t$ the target is
\emph{self-referential}: it is assembled from the network's own directional
derivative, whose accuracy must be assessed separately from the value of
$u_\theta$ at the sampled coordinate. Second, at $h=0$ the correction vanishes and
$\ell_{\mathrm{MF}}$ reduces exactly to the flow-matching loss with the
$\theta$-independent target $\widetilde v_t$. The diagonal $r=t$ is thus the
only place where MeanFlow supervision does not depend on current parameters.
The corresponding marginal-field equation and the baseline's adaptive loss
weighting are given in Appendix~\ref{app:pinn-formulation}.

\subsection{MeanFlow training as temporal collocation}
\label{sec:collocation}

Our MeanFlow baseline \citep{geng2025mean} assigns $75\%$ of the batch to
diagonal samples, drawing $t$ from $\mathrm{LN}(-0.4,1)$ and setting $r=t$.
For the remaining samples, it independently draws $a_1,a_2$ from the same
logit-normal law and sets $(r,t)=(\min(a_1,a_2),\max(a_1,a_2))$.
Two constants are embedded in this
procedure: the fraction of samples placed on the diagonal, fixed at $0.75$, and
the distribution of interval lengths induced by the order statistics of two
logit-normal variables, which places mass on the full range $h\in(0,1)$ from
the first step onward. Neither constant is derived from
Equation~\ref{eq:meanflow-identity}, and neither responds to the fact that
$u_\theta$---and therefore the target in
Equation~\ref{eq:stopped-target}---changes substantially over training.
To make these questions precise we adopt a collocation view. Temporal
coordinates live on the triangle
$\Omega=\{(r,t):0\leq r\leq t\leq 1\}$, whose hypotenuse
$\Gamma_0=\{(t,t)\}$ we call the diagonal. Training draws coordinates from a
step-dependent measure $P_k$ on $\Omega$ and averages
Equation~\ref{eq:stopped-target}. We factor that measure as
\begin{equation}
P_k \;=\; \rho_k\,\bigl(p_t\otimes\delta_{\{r=t\}}\bigr)
\;+\;(1-\rho_k)\,q_k(r,t),
\label{eq:collocation-measure}
\end{equation}
where $p_t$ is the configured time sampler, $\rho_k\in[0,1]$ is the diagonal
fraction at step $k$, and $q_k$ is a distribution supported on the interior
$r<t$. Under this split the loss reads
\begin{equation}
\ell_{\mathrm{MF}}(\theta;x_t,r,t)=
\begin{cases}
\bigl\lVert u_\theta(x_t,t,t)-\widetilde v_t\bigr\rVert_2^2,
& r=t \quad\text{(boundary condition)},\\[4pt]
\bigl\lVert u_\theta(x_t,r,t)-\operatorname{sg}\!\bigl[\widetilde v_t
 -h\,\widetilde{\mathcal D}_t u_\theta(x_t,r,t)\bigr]\bigr\rVert_2^2,
& r<t \quad\text{(interior residual)},
\end{cases}
\label{eq:stopped-pinn-loss}
\end{equation}
and the batch objective is a single mean of Equation~\ref{eq:stopped-pinn-loss}
over $B$ coordinates drawn from $P_k$. This is a renaming, not a new objective:
substituting $\rho_k=0.75$ and the order-statistic law $q^{\mathrm{mm}}$ (for
min--max) recovers standard MeanFlow exactly. The fraction $\rho_k$ determines
sample counts, rather than weighting two separately averaged losses. The boundary condition on
$\Gamma_0$ and the differential constraint on $r<t$ mirror the two collocation
types of a PINN.

\paragraph{Where the PINN analogy stops.}
The split in Equation~\ref{eq:stopped-pinn-loss} is structurally PINN-like, but
it is not a physical boundary-value problem. A deterministic PINN is given its
boundary value, whereas the diagonal MeanFlow target $\widetilde v_t$ is a
sample whose conditional expectation is the unknown marginal velocity.
Consequently, even an expressive predictor faces conditional regression noise
on the diagonal. Interior targets add a second source of error because their
stopped target uses the current network derivative. The analogy therefore
motivates how we organize and schedule the two collocation types; it does not
imply that the diagonal can be solved exactly or that PINN convergence results
transfer to generative training.

\subsection{What the two constants control}
\label{sec:handles}

The collocation view names the two constants but does not yet say what they
act on.
Let $a_\theta=u_\theta(x_t,r,t)-\widetilde v_t$ and
$b_\theta=\operatorname{sg}[\widetilde{\mathcal D}_tu_\theta(x_t,r,t)]$, so that
the interior branch is $\lVert a_\theta+h\,b_\theta\rVert_2^2$. Expanding,
\begin{equation}
\ell_{\mathrm{MF}}
=\underbrace{\lVert a_\theta\rVert_2^2}_{\mathcal L_{\mathrm{TFM}}}
\;+\;\underbrace{2h\,\langle a_\theta,b_\theta\rangle}_{\mathcal L_{\mathrm{TC}}}
\;+\;\underbrace{h^2\lVert b_\theta\rVert_2^2}_{\text{no gradient}},
\qquad
g_{\mathrm{TC}}=2h\,(\nabla_\theta a_\theta)^\top b_\theta .
\label{eq:loss-decomposition}
\end{equation}
This is the trajectory-flow-matching / trajectory-consistency decomposition of
\citet{zhang2025alphaflow}, written per coordinate. The stopped-target gradient
has an explicit factor of $h$:
\begin{equation}
\frac{\lVert g_{\mathrm{TC}}\rVert}{\lVert g_{\mathrm{TFM}}\rVert}
= h\cdot\frac{\lVert(\nabla_\theta a_\theta)^\top b_\theta\rVert}
              {\lVert(\nabla_\theta a_\theta)^\top a_\theta\rVert}
\label{eq:norm-ratio-prediction}
\end{equation}
The remaining factor also depends on the sampled coordinate, so this is not
a monotonicity claim across intervals. The fraction $\rho_k$ controls how many
coordinates carry the consistency term; $h$ enters its per-sample scale.
The collocation view therefore shows that the two constants act on different
quantities and can be varied independently; it does not prescribe whether
either should vary during training.

\subsection{The conflict is not stationary}
\label{sec:nonstationary}

\begin{figure}[t]
  \centering
  \includegraphics[width=0.98\linewidth]{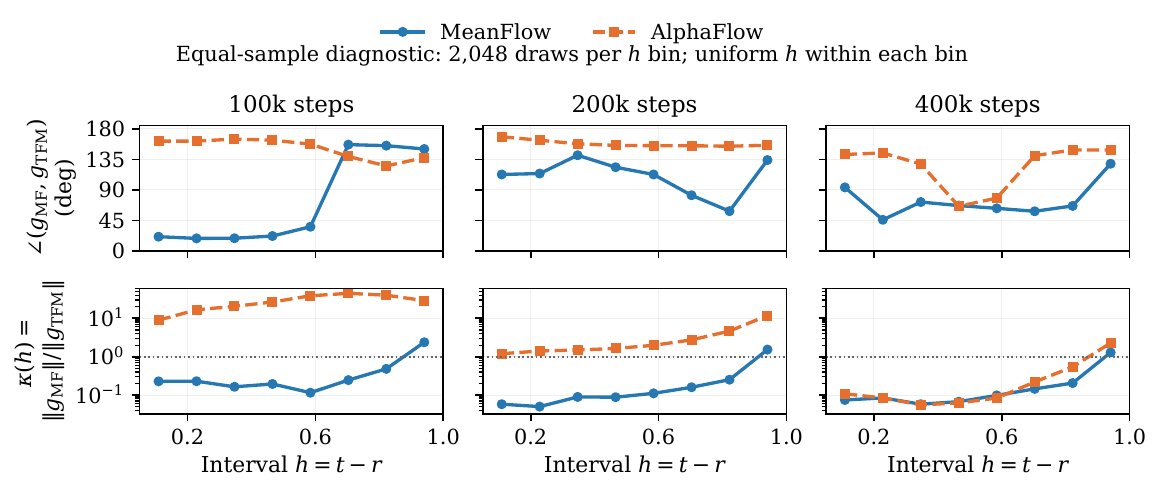}
  \caption{Full-gradient diagnostics for MeanFlow and AlphaFlow at
  100k, 200k, and 400k. Top: $\angle(g_{\mathrm{MF}},g_{\mathrm{TFM}})$ in degrees;
  bottom: $\kappa=\lVert g_{\mathrm{MF}}\rVert/\lVert g_{\mathrm{TFM}}\rVert$,
  unlike the component ratio $q$ in Figure~\ref{fig:three-way-gradient-path}.}
  \label{fig:gradient-conflict-by-interval}
\end{figure}

Equation~\ref{eq:loss-decomposition} fixes the structure of the conflict but
not its magnitude, which depends on $u_\theta$. We therefore probe it directly
on raw DiT-B/2 checkpoints of MeanFlow and AlphaFlow at 100k, 200k and 400k
steps (Figure~\ref{fig:gradient-conflict-by-interval}). Standard MeanFlow keeps
$\rho_k=0.75$ throughout; the sigmoid AlphaFlow run keeps $\rho_k=0.5$, where
``sigmoid'' refers to annealing its target parameter $\alpha$ from $1$ to $0$
rather than to its flow-matching ratio. For each of eight
interval bins we measure the angle between $g_{\mathrm{MF}}=g_{\mathrm{TFM}}+
g_{\mathrm{TC}}$ and $g_{\mathrm{TFM}}$, and the norm ratio
$\kappa(h)=\lVert g_{\mathrm{MF}}\rVert/\lVert g_{\mathrm{TFM}}\rVert$, using
2{,}048 samples in each of eight equal-width bins over $h\in[0.05,1]$.
For nonzero component and full gradients, let
$c=\cos(g_{\mathrm{TFM}},g_{\mathrm{TC}})$ and
$q=\lVert g_{\mathrm{TC}}\rVert/\lVert g_{\mathrm{TFM}}\rVert$.
Then $\kappa=\sqrt{1+q^2+2qc}$ and
$\cos\angle(g_{\mathrm{MF}},g_{\mathrm{TFM}})=(1+qc)/\kappa$.
Figure~\ref{fig:gradient-conflict-by-interval} reports the resulting full-gradient
metrics; Figure~\ref{fig:three-way-gradient-path} reports bin averages of $c$ and
$q$, rather than $\kappa$. Appendix~\ref{app:gradient-probe} gives the probe protocol.
Both direction and scale vary with
interval length and with training progress. At 100k, standard MeanFlow has a
small full-gradient norm over most short and medium intervals, while AlphaFlow
has a much larger norm and a wider angle; the longest intervals are the most
sensitive in both methods. Because the probe uses equal per-bin budgets, it
characterizes the coordinates themselves, not how often a sampler visits them.
We return to the same probe in Section~\ref{sec:gradient-path-results} to
evaluate our schedules.

\FloatBarrier

\paragraph{Design choices.}
These observations motivate three testable choices:
(i) begin with the diagonal's $\theta$-independent target before admitting
interior samples (Section~\ref{sec:ratio-curriculum});
(ii) control interval length independently of the diagonal fraction
(Section~\ref{sec:interval-curriculum}); and (iii) vary both controls to probe
whether adaptation helps as gradient direction and scale change over training
(Section~\ref{sec:joint-schedule}). These choices affect only the sampling
measure; the fixed-horizon control in Section~\ref{sec:ablation} tests a limit
of the joint schedule.

\section{Temporal Supervision Allocation}
\label{sec:method}

TSA replaces the fixed collocation measure with
\begin{equation}
P_k^{\mathrm{TSA}}
=\rho_k\bigl(p_t\otimes\delta_{\{r=t\}}\bigr)
+(1-\rho_k)q^{\mathrm{la}}_{\hmax(k)},
\label{eq:tsa-measure}
\end{equation}
where $\rho_k$ is the diagonal fraction and $q^{\mathrm{la}}_H$ is a
left-anchored interval law with horizon $H$. Relative to MeanFlow
($\rho_k=0.75$, $q^{\mathrm{mm}}$), TSA changes three sampling choices---allocation,
interval parameterization, and horizon---whose effects
Section~\ref{sec:ablation} separates.
The per-sample loss and target remain unchanged.

\subsection{Boundary-to-interior allocation curriculum}
\label{sec:ratio-curriculum}

The diagonal provides the only $\theta$-independent target
(Section~\ref{sec:prelim}). Following the easy-to-hard optimization principle
of \citet{krishnapriyan2021pinn}, we first fit this anchor and progressively
admit self-referential interior constraints. We use
\begin{equation}
\bar\rho_k=\rho_{\mathrm{start}}+
(\rho_{\mathrm{end}}-\rho_{\mathrm{start}})
\sigma\!\left(\gamma
\frac{k-k_{\mathrm{mid}}}{k_{\mathrm{end}}-k_{\mathrm{start}}}\right),
\label{eq:ratio-schedule}
\end{equation}
We implement $\rho_k=1$ when $\bar\rho_k>0.995$ and
$\rho_k=\bar\rho_k$ otherwise, before per-device batch rounding.
The reported run uses
$(\rho_{\mathrm{start}},\rho_{\mathrm{end}})=(1,0.65)$,
$\gamma=20$, and $(k_{\mathrm{start}},k_{\mathrm{mid}},k_{\mathrm{end}})
=(0,200\mathrm{k},400\mathrm{k})$.
Starting at $\rho=1$ separates boundary fitting from interior learning;
ending at $0.65$ retains the anchor while increasing the interior fraction
from MeanFlow's $25\%$ to $35\%$.

For each per-device batch of size $B$, we place $\lfloor B\rho_k\rfloor$
samples on $r=t$, drawing $t$ from the unchanged sampler $p_t$, and draw the
remaining pairs from $q_k$;
Appendix~\ref{app:implementation-correspondence} specifies the realized
fractions and adaptive per-sample loss.

\subsection{Short-to-long interval curriculum}
\label{sec:interval-curriculum}

Interval length controls the explicit factor $h=t-r$ in the consistency
gradient (Equation~\ref{eq:loss-decomposition}). Capping $h$ initially selects
short-interval constraints that approach the diagonal as $h\to0$, then admits
longer intervals toward the inference coordinate $(0,1)$.

Independent variables
$\eta_h,\eta_r\sim\operatorname{LogitNormal}(-0.4,1.0^2)$ define
\begin{equation}
h=\hmax(k)\eta_h,\qquad r=(1-h)\eta_r,\qquad t=r+h.
\label{eq:left-anchor}
\end{equation}
This guarantees $0\leq r\leq t\leq1$ and $h\leq\hmax(k)$.
For any admissible $h$, $r$ ranges over $[0,1-h]$; collectively, the sampler
retains support throughout $t\in[0,1]$. It restricts interval length while
preserving global time coverage.

The joint run uses the quadratic horizon
\begin{equation}
\hmax(k)=
\begin{cases}
0.2, & k\leq100\mathrm{k},\\[2pt]
0.2+0.8\left(\dfrac{k-100\mathrm{k}}{250\mathrm{k}}\right)^2,
&100\mathrm{k}<k<350\mathrm{k},\\[2pt]
1, & k\geq350\mathrm{k}.
\end{cases}
\label{eq:interval-horizon-schedule}
\end{equation}
Its convex shape delays large-$h$ supervision relative to a linear ramp,
motivated by the component-scale imbalance in
Equation~\ref{eq:loss-decomposition}. This is a design choice:
Section~\ref{sec:ablation} compares it with a full horizon, but does not
establish that quadratic growth is optimal.

\paragraph{Interval law versus horizon.}
Even at $\hmax=1$, left anchoring differs from MeanFlow's min--max sampler.
With $2\times10^6$ Monte Carlo draws per law, the mean interior interval is
about $0.42$ versus $0.23$, and the mass near the inference corner
($r<0.1$, $t>0.9$) is $1.8\%$ versus $0.035\%$.
Greater exposure near deployment may explain part of the parameterization
benefit, although the mechanism is not isolated. To separate it from
horizon scheduling, the full-horizon control uses the same $\rho_k$ schedule
and left-anchored law as TSA, with $\hmax\equiv1$.

\subsection{Joint schedule}
\label{sec:joint-schedule}

Figure~\ref{fig:teaser-collocation} shows the joint curriculum, which moves
through boundary fitting ($k\leq100\mathrm{k}$), interior admission
($100\mathrm{k}<k<350\mathrm{k}$), and full-horizon training
($k\geq350\mathrm{k}$). We expect reduced gradient conflict during interior
admission. Once the schedules settle, TSA and the baselines face the same
constraint family with different allocation and interval laws, so the
advantage need not persist (Section~\ref{sec:gradient-path-results}).
A coordinate-level exposure proxy, $S_k=(1-\rho_k)\mathbb{E}_{q_k}[t-r]$,
shows that TSA supplies $0.41\times$ MeanFlow's interior exposure at 200k and
$2.51\times$ at 400k, while its cumulative exposure through 400k is $88.2\%$ of
MeanFlow's (Appendix~\ref{app:schedule-details}). This proxy measures coordinate
exposure rather than loss or gradient strength;
the schedule changes its timing, distribution, and cumulative amount.

\subsection{Design choices, cost, and scope}
\label{sec:design-choices}

\paragraph{Global coverage.}
Restricting $t$ may induce covariate shift as the shared generative network
moves between marginals; this explanation remains a hypothesis.
Left anchoring instead restricts $h$ while preserving global time coverage.

\paragraph{Resampling and target construction.}
For a fixed per-sample loss, changing the sampling density can be represented
by density-ratio loss weights in expectation, provided the proposal covers
the required support \citep{kingma2023understanding}.
This does not imply identical estimator variance or identical behavior when
weights are inserted before adaptive loss normalization
\citep{kim2026understanding}. TSA changes the training measure without
importance compensation, retaining the baseline's single adaptively weighted
batch mean (Appendix~\ref{app:implementation-correspondence}).
It schedules sample counts and interval support while preserving the
per-sample loss; we do not claim resampling is universally better than weighting.
AlphaFlow instead anneals the target between finite-difference and continuous
derivatives \citep{zhang2025alphaflow}. We fix $\alpha=0$;
combining the two interventions is possible but untested, and their gains
need not be additive.

\begin{figure}[!ht]
\centering
\begin{minipage}[t]{0.40\linewidth}
\vspace{0pt}
\paragraph{Training step.}
Algorithm~\ref{alg:tsa} adds two schedules and draws pairs $(r,t)$.
It adds no model parameters or inference steps. We still compute JVPs on
the full batch, so we claim no faster training. Here LN is the baseline
logit-normal law; $\widetilde v_i$ is the conditional velocity, and
$\ell_{\rm base}$ is the baseline loss.

\paragraph{Scope.}
TSA changes the training path but leaves gradient conflict in place.
We test this with different interval laws and inference samplers; the
horizon rule does not improve every protocol.
\end{minipage}\hfill
\begin{minipage}[t]{0.57\linewidth}
\vspace{0pt}
\refstepcounter{algorithm}
\algorithmcaption{EbbFlow: TSA training step.}
\label{alg:tsa}
\par\vspace{0.5em}
\begin{tabular}{@{}l@{}}
$H\gets h_{\max}(k),\quad n\gets\lfloor B\rho_k\rfloor,\quad L\gets0$\\
\textbf{for} $i=1,\ldots,B$:\\
\quad\textbf{if} $i\leq n$: $t\sim p_t,\quad r\gets t$\\
\quad\textbf{else}:\\
\qquad $a,b\stackrel{\rm iid}{\sim}\mathrm{LN}(-0.4,1)$\\
\qquad $h\gets Ha,\quad r\gets(1-h)b,\quad t\gets r+h$\\
\quad $x_t\gets(1-t)x_i+t\epsilon_i$\\
\quad $\widetilde v_i\gets\mathrm{baseline\_velocity}(x_i,\epsilon_i,t)$\\
\quad $u\gets u_\theta(x_t,r,t)$\\
\quad $g\gets\operatorname{JVP}(u_\theta;(x_t,r,t),(\widetilde v_i,0,1))$\\
\quad $y\gets\operatorname{stopgrad}(\widetilde v_i-(t-r)g)$\\
\quad $L\gets L+\ell_{\rm base}(u,y)/B$\\
$\operatorname{update}(\theta,L)$
\end{tabular}
\par\vspace{0.2em}\hrule
\end{minipage}
\end{figure}
\FloatBarrier
\section{Experiments}
\label{sec:experiments}
\suppressfloats[t]

\subsection{Evaluation protocol}
\label{sec:setup}

All matched quality runs use class-conditional latent ImageNet
\citep{deng2009imagenet} $256\times256$ generation with DiT-B/2
\citep[131M parameters;][]{peebles2023scalable}, a global
batch of 256, and 400k updates. MeanFlow keeps $\rho_k=0.75$ and uses
min--max intervals $q^{\mathrm{mm}}$. AlphaFlow uses sigmoid target
annealing with $\rho_k=0.5$. EbbFlow retains the MeanFlow target
($\alpha=0$) and uses Equation~\ref{eq:tsa-measure}. The runs share
the optimizer, EMA rule, checkpoint selection, random class sampling,
and evaluation code. All matched runs are class-conditional but use no
classifier-free guidance \citep{ho2022classifier}: the training target uses
the conditional velocity, and the inference guidance scale is one.
We report FID \citep{heusel2017gans} and DINO-FID (FDD) from 50k
generated images at one and two network evaluations (NFE). FDD uses
DINOv2-ViT-L/14 features \citep{stein2023exposing}. The
two-NFE results in Table~\ref{tab:main} use ODE sampling.

Each matched training configuration has one seed. The AlphaFlow row
averages two independent 50k-sample evaluations of the same checkpoint;
it is not a two-seed training average. Published results with different
backbones, budgets, or protocols
are not directly comparable.

\begin{table*}[htbp]
\centering
\begin{minipage}[b]{0.30\textwidth}
\raggedright
\caption{Matched ImageNet $256\times256$ results (\mbox{DiT-B/2},
400k updates; \mbox{50k-sample} FID). Two-NFE uses ODE sampling.
Full metrics: \mbox{Table~\ref{tab:main-full}}.}
\label{tab:main}
\vspace{\baselineskip}
\centering
\setlength{\tabcolsep}{3pt}
\begin{tabular}[b]{@{}lcc@{}}
\toprule
Method & 1-NFE & 2-NFE\\
 & & ODE\\
\midrule
MeanFlow & 44.82 & 39.86\\
AlphaFlow & 41.89 & 37.67\\
EbbFlow & \textbf{39.65} & \textbf{36.02}\\
\bottomrule
\end{tabular}
\par
\end{minipage}
\hfill
\begin{minipage}[b]{0.67\textwidth}
\raggedright
\caption{DiT-B/2 ablations at 400k updates. MM: min--max;
LA: left anchor; Q: quadratic horizon (otherwise full).
Full metrics: Table~\ref{tab:ablation-full}.}
\label{tab:ablation}
\vspace{\baselineskip}
\centering
\setlength{\tabcolsep}{3pt}
\begin{tabular}[b]{@{}lllcc@{}}
\toprule
 & $\rho_k$ & Sampler & 1-NFE & 2-NFE\\
 & & & & ODE\\
\midrule
(a) & $0.75$ & MM & 44.82 & 39.86\\
(b) & $0.82525$ & MM & 44.13 & 38.77\\
(c) & $1\to0.65$ & MM & 41.80 & 37.22\\
(d) & $0.75$ & LA-Q & 40.78 & 36.92\\
(e) & $1\to0.65$ & LA & \textbf{38.85} & 36.25\\
(f) & $1\to0.65$ & LA-Q & 39.65 & \textbf{36.02}\\
\bottomrule
\end{tabular}
\par
\end{minipage}
\par\vspace{\baselineskip}
\end{table*}

\subsection{Matched ImageNet results}
\label{sec:main}

Under this protocol, EbbFlow lowers FID relative to matched MeanFlow by
5.17 points (11.5\%) at one NFE and 3.84 points (9.6\%) at two NFE.
Relative to AlphaFlow, the reductions are 2.24 and 1.65 points.
EbbFlow has the lowest two-NFE FDD, but its one-NFE FDD is 9.1 points
\emph{higher} than AlphaFlow's (Table~\ref{tab:main-full}). The metrics therefore disagree on
the one-NFE ordering of those two methods. These single-seed
comparisons do not estimate training-seed variability.

\subsection{Attribution of temporal-coordinate choices}
\label{sec:ablation}

Table~\ref{tab:ablation} separates allocation, interval law, and horizon.
The full-horizon left-anchor control shares EbbFlow's allocation schedule
and interval law, isolating horizon expansion.

\paragraph{Allocation.}
With min--max intervals fixed, scheduling the allocation improves
one-NFE FID by 3.02 points over MeanFlow (a$\to$c) and by 2.33
points over the static-mixture control (b$\to$c). With the
quadratic left-anchor horizon fixed, the allocation schedule
improves one-NFE FID by 1.13 points (d$\to$f). The static control
suggests that temporal allocation matters beyond an approximately matched
average mixture. Per-device rounding gives slightly different realized
diagonal fractions (about $82.03\%$ static versus $82.29\%$ scheduled),
so this is not an exact discrete-budget control.

\paragraph{Interval law and horizon.}
At the same scheduled allocation, replacing min--max intervals (c)
with full-horizon left-anchored intervals (e) lowers one-NFE FID by
2.95 points and two-NFE ODE FID by 0.97 points. This isolates the
interval-law effect under scheduled allocation, not under fixed
$\rho_k=0.75$. Adding the quadratic horizon (e$\to$f)
\emph{raises} one-NFE FID by 0.80 points and consistency FID by
0.60 points (Table~\ref{tab:ablation-full}), but \emph{lowers} two-NFE ODE FID by 0.23 points.
Thus the horizon schedule is not uniformly beneficial. A
linear-horizon control and multiple seeds would be needed to test
whether its shape or the small two-NFE difference is robust.

\paragraph{Sampler dependence.}
\label{sec:sampling}
With each checkpoint held fixed, two-NFE ODE sampling beats consistency
sampling for every $\alpha=0$ variant, by 3.12 to 6.01 FID
(Table~\ref{tab:ablation-full}). Our two-NFE claims are therefore specific to
ODE sampling and do not establish which sampler suits AlphaFlow's target.

\subsection{Optimization-path comparison}
\label{sec:gradient-path-results}

We apply the equal-bin probe of Section~\ref{sec:nonstationary} to raw
100k, 200k, 300k, and 400k checkpoints of all three methods (FP32,
evaluation mode), using 2{,}048 samples per bin. It diagnoses coordinate-level
gradient geometry rather than the training sampler's visit frequency.

At 100k, EbbFlow is effectively in its pure boundary-fitting phase.
The probe nevertheless includes untrained interior coordinates,
yielding a less-negative mean component cosine ($-0.716$) and a large
component norm ratio $q$ (48.9). This is task separation, not balanced joint
optimization. At 200k, the mean $q$ falls to 1.25, near
MeanFlow's 1.18 and below AlphaFlow's 4.31. EbbFlow's mean cosine
is also less negative ($-0.908$ versus $-0.979$ and $-0.960$).
A smaller directional advantage persists at 300k. By 400k,
all three mean cosines are about $-0.98$, and the mean norm ratios
lie between 1.10 and 1.38. These measurements suggest a mid-training
difference followed by similar bin-averaged diagnostics at 400k. This
post-hoc probe cannot by itself establish that the path change
caused the final FID differences.

\begin{figure}[!ht]
  \centering
  \includegraphics[width=0.98\linewidth]{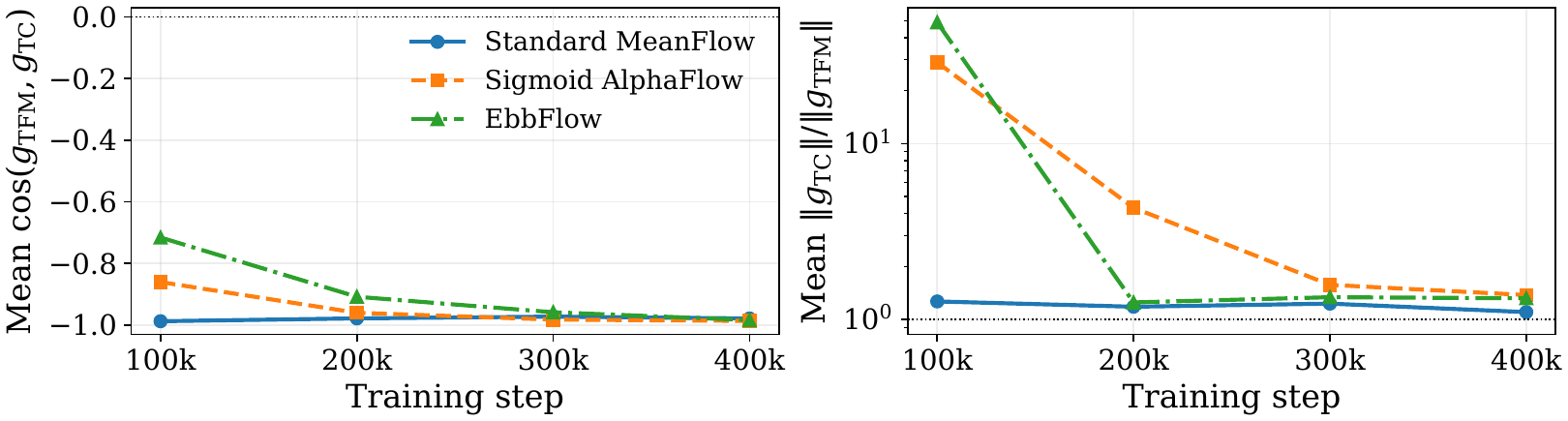}
  \caption{Component-gradient diagnostics at 100k--400k.
  Left: mean $\cos(g_{\mathrm{TFM}},g_{\mathrm{TC}})$;
  right: mean $q=\lVert g_{\mathrm{TC}}\rVert/\lVert g_{\mathrm{TFM}}\rVert$.
  Each point averages eight interval bins.}
  \label{fig:three-way-gradient-path}
\end{figure}

\FloatBarrier
\section{Discussion and Conclusion}
\label{sec:discussion}\label{sec:conclusion}

We presented \method{}, which schedules diagonal allocation and interval
geometry in MeanFlow's temporal sampler. In matched single-seed DiT-B/2 runs
at 400k updates, it improves MeanFlow's FID from 44.82 to 39.65 at 1-NFE
and from 39.86 to 36.02 at 2-NFE ODE. At 200k, gradient probes suggest less
component-gradient opposition than both baselines and better scale balance
than AlphaFlow; by 400k, bin-averaged diagnostics are similar across methods.
Ablations support benefits from allocation scheduling and left-anchored
intervals, while the full-horizon left-anchor control outperforms the joint
curriculum for two of three samplers. The evidence supports temporal sampling
as a complementary training control and a 2-NFE benefit from the joint
curriculum. Multi-seed matched runs remain necessary to establish reliability.

\subsection*{AI use statement}
Generative AI tools assisted with finding and checking related literature,
checking and clarifying parts of the mathematical analysis, reviewing the
experimental methodology and interpretation of results, drafting and editing
manuscript text, preparing code and figures, and checking formatting.
The authors reviewed the AI-assisted
analysis, text, code, and figures and take responsibility for the final content,
claims, and artifacts in this submission.

\subsection*{Reproducibility statement}
Section~\ref{sec:method} and Algorithm~\ref{alg:tsa} specify TSA;
Appendices~\ref{app:implementation-correspondence}--\ref{app:gradient-probe},
\ref{app:schedule-details}, \ref{app:complete-metrics},
and~\ref{app:experimental-configurations} describe implementation details,
the diagnostic protocol, schedules, complete evaluation metrics, and
experimental configurations.
Section~\ref{sec:experiments} and Appendix~\ref{app:experimental-configurations}
state the training-seed count, evaluation sample counts, checkpoint selection,
sampling modes, batch sizes, and guidance settings.

% Funding acknowledgments are omitted from the anonymous review version.
\ificlrfinal
\subsubsection*{Acknowledgments}
This work was also supported by the Guangxi Science and Technology Major
Program (Grant No.\ AA24206054).
\fi

% Long source URLs create excessive interword spacing in justified entries.
% Keep the official BibTeX style and the complete, clickable URLs.
\begingroup
\raggedright
\bibliography{references}
\bibliographystyle{iclr2027_conference}
\endgroup

\appendix

\section{Detailed PINN-like formulation}
\label{app:pinn-formulation}

\subsection{Characteristic constraint and diagonal trace}

Let $x_0$ denote a data sample, $\epsilon$ a noise sample, and
$x_t=(1-t)x_0+t\epsilon$. A model $u_\theta(x_t,r,t)$ predicts transport over
an interval $0\leq r\leq t\leq1$. Holding $r$ fixed, define the characteristic
derivative
\begin{equation}
\mathcal{D}_t\triangleq\partial_t+v(x_t,t)^\top\nabla_{x_t}.
\label{eq:characteristic-derivative}
\end{equation}
The ideal MeanFlow identity in Equation~\ref{eq:meanflow-identity} motivates
the interior constraint
\begin{equation}
\mathcal{R}_\theta(x_t,r,t)
\triangleq u_\theta(x_t,r,t)-v(x_t,t)
+(t-r)\mathcal{D}_t u_\theta(x_t,r,t)=0,
\qquad r<t.
\label{eq:meanflow-pde-residual}
\end{equation}
On the diagonal boundary $\Gamma_0=\{r=t\}$, the coefficient of the
derivative term vanishes, yielding
\begin{equation}
\mathcal{B}_\theta(x_t,t)
\triangleq u_\theta(x_t,t,t)-v(x_t,t)=0.
\label{eq:meanflow-boundary}
\end{equation}
The same structure has an initial-value representation. Define the displacement
$S_\theta(x_t,r,t)=(t-r)u_\theta(x_t,r,t)$. By the product rule,
Equation~\ref{eq:meanflow-pde-residual} becomes
\begin{equation}
\mathcal{D}_tS_\theta(x_t,r,t)=v(x_t,t),
\qquad S_\theta(x_r,r,r)=0.
\label{eq:displacement-initial-trace}
\end{equation}
The zero-displacement trace holds for any bounded $u_\theta$; by itself it
does not enforce Equation~\ref{eq:meanflow-boundary}. Together with the interior
constraint and sufficient regularity, it recovers the diagonal relation.

This is a PINN-like optimization view rather than a conventional deterministic
PINN. The states $x_t$ are sampled from data--noise interpolations, and the
unknown marginal velocity $v(x_t,t)$ is supervised using conditional
velocities. The view is nevertheless useful because it separates diagonal
boundary collocation from interior differential-constraint collocation, which
are precisely the two types of samples controlled by our schedules.

\subsection{Stopped target and implementation correspondence}
\label{app:implementation-correspondence}

We retain the continuous MeanFlow target and modify only temporal sampling.
In the AlphaFlow implementation, this corresponds to fixing $\alpha=0$, so
the finite-difference branch is inactive. Let $\widetilde v_t$ denote the
training velocity: $\widetilde v_t=\epsilon-x_0$ without guidance, or the
velocity returned by the baseline's training-time guidance construction. For
each sample, the implementation computes
\begin{align}
g_\theta(x_t,r,t)
&=\operatorname{JVP}\bigl(u_\theta,(x_t,r,t),
  (\widetilde v_t,0,1)\bigr),\\
y_\theta(x_t,r,t)
&=\operatorname{sg}\bigl[\widetilde v_t-(t-r)g_\theta(x_t,r,t)\bigr],
\label{eq:interval-target}
\end{align}
where $\operatorname{sg}$ stops gradients through the target. When $r=t$,
the JVP contribution vanishes and $y_\theta=\widetilde v_t$. For a
$d$-dimensional output, the baseline's adaptively weighted per-sample loss is
\begin{equation}
e_i=\frac{1}{d}\left\lVert
u_\theta(x_{t_i},r_i,t_i)-y_\theta(x_{t_i},r_i,t_i)
\right\rVert_2^2,
\qquad
\ell_i=\frac{e_i}{\operatorname{sg}(e_i)+\varepsilon_{\mathrm w}},
\quad \varepsilon_{\mathrm w}=10^{-3}.
\label{eq:implemented-loss}
\end{equation}
The conditional tangent and stopped target form a stochastic training surrogate
for the ideal marginal-field constraint. Per sample, the conditional velocity
need not equal the marginal velocity; stopping the JVP target also gives a
different parameter derivative from direct minimization of
$\lVert\mathcal R_\theta\rVert^2$.

For a batch of size $B$, the code assigns
\begin{equation}
B_{\mathrm{FM}}=\lfloor B\rho_k\rfloor,
\qquad B_{\mathrm{int}}=B-B_{\mathrm{FM}},
\qquad \widehat\rho_k=B_{\mathrm{FM}}/B,
\label{eq:finite-batch-allocation}
\end{equation}
and averages $\ell_i$ over all samples. Hence $\rho_k$ changes sample counts,
not an external coefficient applied after separately averaging two losses.
Diagonal times follow the fixed configured logit-normal sampler, while
off-diagonal pairs follow $q_k(r,t)$. Table~\ref{tab:code-correspondence}
records the notation used by the implementation.

\begin{table}[htbp]
\caption{Correspondence between the paper notation and the AlphaFlow training
implementation.}
\label{tab:code-correspondence}
\vspace{\baselineskip}
\centering
\begin{tabular}{@{}ll@{}}
\toprule
Paper notation & Implementation quantity \\
\midrule
$r$ & \texttt{t\_next} or \texttt{sigma\_next} \\
$t$ & \texttt{t} or \texttt{sigma} \\
$x_t$ & \texttt{x\_t} \\
$\widetilde v_t$ & \texttt{velocity\_cfg} \\
$u_\theta(x_t,r,t)$ & \texttt{pred\_mean\_velocity} \\
$g_\theta$ & \texttt{videos\_dudt\_mf} \\
$y_\theta$ & output of \texttt{\_compute\_mean\_velocity\_c} \\
$\rho_k$ & \texttt{ratio\_fm} \\
\bottomrule
\end{tabular}
\end{table}

\subsection{Gradient-probe protocol}
\label{app:gradient-probe}

Both diagnostic figures use the continuous MeanFlow target ($\alpha=0$),
including for AlphaFlow checkpoints; they do not measure AlphaFlow's annealed
training target. Raw checkpoints run in FP32 and evaluation mode.
Eight equal-width bins partition $h\in[0.05,1]$; each uses 2{,}048 samples,
with $h$ uniform within the bin and $r$ uniform on $[0,1-h]$.
The seed is 20260907. Sample indices, noise, and class conditions match across
methods, but latent banks are generated separately and need not be identical.
Each method reuses its bank across bins and checkpoints.
Gradients of the two components include the common stopped adaptive weight
from Equation~\ref{eq:implemented-loss} and are averaged over samples before
computing direction and norm statistics. No sampling-frequency weights enter.
Figure~\ref{fig:gradient-conflict-by-interval} transforms these statistics to
full-gradient metrics in each bin; Figure~\ref{fig:three-way-gradient-path}
takes arithmetic means of the component metrics over bins. Curves are point
estimates, without uncertainty estimates across training seeds.

\FloatBarrier
\section{TSA schedule details}
\label{app:schedule-details}

\paragraph{Schedule values.}
At 100k, 200k, 300k, and 400k updates, the clamped diagonal fractions before
per-device batch rounding are
$1$, $0.825$, $0.652$, and $0.650$, and the horizons are $0.2$, $0.328$,
$0.712$, and $1.0$. During boundary fitting ($k\leq100\mathrm{k}$) the horizon
is inactive, so a probe of the still-untrained interior coordinates can have
a large norm ratio.

\begin{figure}[!ht]
  \centering
  \includegraphics[width=0.98\linewidth]{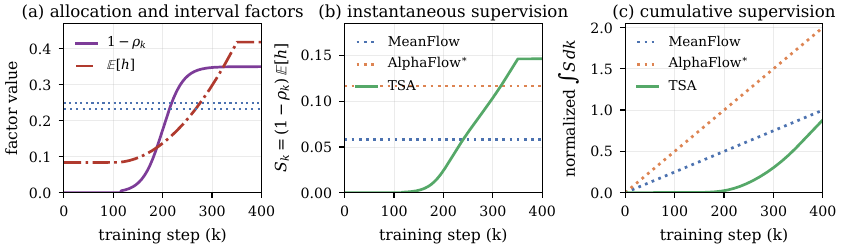}
  \caption{Sampler-based exposure before batch rounding: (a) interior fraction and mean interval;
  (b) their product $S_k$; (c) cumulative exposure relative to MeanFlow's
  400k total. AlphaFlow$^\ast$ denotes coordinate exposure, not gradient strength.}
  \label{fig:effective-supervision}
\end{figure}

\paragraph{Effective supervision.}
We summarize the coordinate-level exposure to the consistency component by
\begin{equation}
S_k=(1-\rho_k)\mathbb{E}_{(r,t)\sim q_k}[t-r].
\label{eq:effective-supervision}
\end{equation}
This pre-rounding proxy combines interior frequency with mean interval length;
it is neither a measured loss nor a gradient norm.
TSA supplies $0.41\times$ MeanFlow's instantaneous exposure at 200k and
$2.51\times$ at 400k, with cumulative exposure of $88.2\%$ through 400k
(Figure~\ref{fig:effective-supervision}).
The late ratio combines $1.4\times$ more interior samples with an approximately
$1.8\times$ longer mean interval.

\FloatBarrier
\section{Complete generation metrics}
\label{app:complete-metrics}

Tables~\ref{tab:main-full} and~\ref{tab:ablation-full} retain the full metrics
corresponding to the compact main-text tables. Each configuration has one
training seed; all evaluations use 50k generated samples.

\begin{table}[!ht]
\caption{Complete matched ImageNet $256\times256$ results with DiT-B/2 at
400k updates. AlphaFlow averages two evaluations of the same checkpoint.
Lower FID and FDD are better; two-NFE results use ODE sampling.}
\label{tab:main-full}
\centering
\vspace{\baselineskip}
\begin{tabular}{@{}lcccc@{}}
\toprule
 & \multicolumn{2}{c}{1-NFE} & \multicolumn{2}{c}{2-NFE ODE}\\
\cmidrule(lr){2-3}\cmidrule(lr){4-5}
Method & FID & FDD & FID & FDD\\
\midrule
MeanFlow & 44.82 & 843.8 & 39.86 & 805.7\\
AlphaFlow & 41.89 & \textbf{787.8} & 37.67 & 772.2\\
EbbFlow & \textbf{39.65} & 796.9 & \textbf{36.02} & \textbf{768.2}\\
\bottomrule
\end{tabular}
\par\vspace{\baselineskip}
\end{table}

\begin{table}[!ht]
\caption{Complete DiT-B/2 ablations at 400k updates, measured by
50k-sample FID. MM denotes min--max intervals and LA denotes left anchoring.
Static $\rho=0.82525$ approximates the time-averaged allocation of
$1\to0.65$. Cons. denotes two-NFE consistency sampling.}
\label{tab:ablation-full}
\centering
\vspace{\baselineskip}
\begin{tabular}{@{}lll lccc@{}}
\toprule
 & $\rho_k$ & Law & Horizon & 1-NFE & 2-NFE ODE & 2-NFE cons.\\
\midrule
(a) & $0.75$ & MM & full & 44.82 & 39.86 & 45.51\\
(b) & $0.82525$ & MM & full & 44.13 & 38.77 & 44.78\\
(c) & $1\to0.65$ & MM & full & 41.80 & 37.22 & 42.35\\
(d) & $0.75$ & LA & quadratic & 40.78 & 36.92 & 41.67\\
(e) & $1\to0.65$ & LA & full & \textbf{38.85} & 36.25 & \textbf{39.37}\\
(f) & $1\to0.65$ & LA & quadratic & 39.65 & \textbf{36.02} & 39.97\\
\bottomrule
\end{tabular}
\par\vspace{\baselineskip}
\end{table}

\clearpage
\raggedbottom
\section{Detailed experimental configurations}
\label{app:experimental-configurations}

Table~\ref{tab:shared-configurations} lists the shared architecture, training,
and evaluation settings of the matched DiT-B/2 experiments.
Table~\ref{tab:temporal-configurations} specifies the temporal samplers.
These settings describe the matched DiT-B/2 runs rather than the larger-model
settings reported in prior work.

\par\vspace{4pt}
\begin{minipage}{\linewidth}
\makeatletter\def\@captype{table}\makeatother
\caption{Shared configurations for the matched ImageNet $256\times256$
experiments. Parameter count refers to the trainable DiT backbone and excludes
the frozen autoencoder. The learning rate includes a 10k-update warm-up.}
\label{tab:shared-configurations}
\vspace{\baselineskip}
\centering
\begin{tabular}{@{}ll@{}}
\toprule
Configuration & Value\\
\midrule
\multicolumn{2}{@{}l}{\emph{Network architecture and data}}\\
Backbone & DiT-B/2\\
Backbone parameters (M) & approximately 131\\
Transformer depth / hidden dimension & $12$ / $768$\\
Attention heads / latent patch size & $12$ / $2\times2$\\
MLP expansion ratio & $4$\\
Image resolution / latent resolution & $256\times256$ / $32\times32$\\
Latent channels / conditioning classes & $4$ / $1{,}000$\\
Frozen autoencoder & sd-vae-ft-ema\\
\midrule
\multicolumn{2}{@{}l}{\emph{Training hyperparameters}}\\
Training updates / fine-tuning & $400\mathrm{k}$ / none\\
Global batch size / training seed & $256$ / $1$\\
Optimizer & Adam\\
Peak and post-warm-up learning rate & $10^{-4}$\\
Learning-rate schedule & 10k linear warm-up, then constant\\
Adam $(\beta_1,\beta_2)$ / $\epsilon$ & $(0.9,0.95)$ / $10^{-8}$\\
Weight decay / dropout & $0$ / $0$\\
Gradient clipping norm & $16$\\
Training autocast precision & bfloat16\\
EMA half-life (updates) / ramp-up ratio & $6{,}931$ / $0.05$\\
EMA update frequency & every update\\
Adaptive loss denominator offset & $10^{-3}$\\
Guidance $(\omega_{\mathrm{CFG}},\kappa_{\mathrm{CFG}})$ & $(1,0)$ (disabled)\\
Class-label dropout probability & $0$\\
\midrule
\multicolumn{2}{@{}l}{\emph{Generation and evaluation}}\\
Generation checkpoint & 400k-update EMA\\
Generated samples per metric evaluation & $50{,}000$\\
Noise standard deviation / time shift & $1$ / $1$\\
Inference guidance scale & $1$\\
One-NFE time grid & $1\to0$\\
Two-NFE time grid & $1\to0.5\to0$\\
Two-NFE samplers & ODE and consistency\\
Feature metrics & FID; DINO-FID (DINOv2-ViT-L/14)\\
\bottomrule
\end{tabular}
\end{minipage}
\par\vspace{4pt}

For EbbFlow and ablations (b)--(e) in Table~\ref{tab:ablation}, two GPUs
process 128 samples each; the matched MeanFlow run uses four GPUs with
64 samples each. These runs retain the same global batch of 256.
All ablations use the shared settings in Table~\ref{tab:shared-configurations}
and $\alpha=0$. Their changes to allocation, interval law, and horizon are
listed in Table~\ref{tab:ablation-full}; active allocation and horizon schedules
use the EbbFlow parameters in Table~\ref{tab:temporal-configurations}.
The static allocation $0.82525$ approximates the time-averaged scheduled
allocation. Each training configuration has one seed. The AlphaFlow main-table
row averages two metric evaluations of the same checkpoint.
\par\vspace{4pt}
\begin{minipage}{\linewidth}
\makeatletter\def\@captype{table}\makeatother
\caption{Temporal-sampler configurations for the matched baselines and
EbbFlow. LN denotes $\operatorname{LogitNormal}(-0.4,1.0^2)$; MM denotes
min--max ordering of two independent draws, and LA denotes
Equation~\ref{eq:left-anchor}. Dashes indicate an inactive schedule.}
\label{tab:temporal-configurations}
\vspace{\baselineskip}
\centering
\setlength{\tabcolsep}{5pt}
\begin{tabular}{@{}lccc@{}}
\toprule
Configuration & MeanFlow & AlphaFlow & EbbFlow\\
\midrule
Diagonal timestep law & LN & LN & LN\\
Interior coordinate law & MM (LN) & MM (LN) & LA (LN)\\
Diagonal fraction $\rho_k$ & $0.75$ & $0.50$ & $1\to0.65$\\
Allocation schedule & constant & constant & sigmoid\\
Allocation start / end (updates) & --- & --- & $0$ / $400\mathrm{k}$\\
Allocation steepness $\gamma_\rho$ & --- & --- & $20$\\
Allocation endpoint clamp & --- & --- & $0.005$\\
Interval horizon $h_{\max}$ & $1$ & $1$ & $0.2\to1$\\
Horizon start / end (updates) & --- & --- & $100\mathrm{k}$ / $350\mathrm{k}$\\
Horizon power & --- & --- & $2$\\
Target parameter $\alpha$ & $0$ & $1\to0$ & $0$\\
Target schedule & constant & sigmoid & constant\\
Target start / end (updates) & --- & $0$ / $400\mathrm{k}$ & ---\\
Target steepness / endpoint clamp & --- & $25$ / $0.005$ & ---\\
\bottomrule
\end{tabular}
\end{minipage}
\par\medskip

\clearpage
\flushbottom
\section{Qualitative samples}
\label{app:qualitative-samples}

\begin{figure}[!ht]
  \centering
  \includegraphics[width=0.98\linewidth]{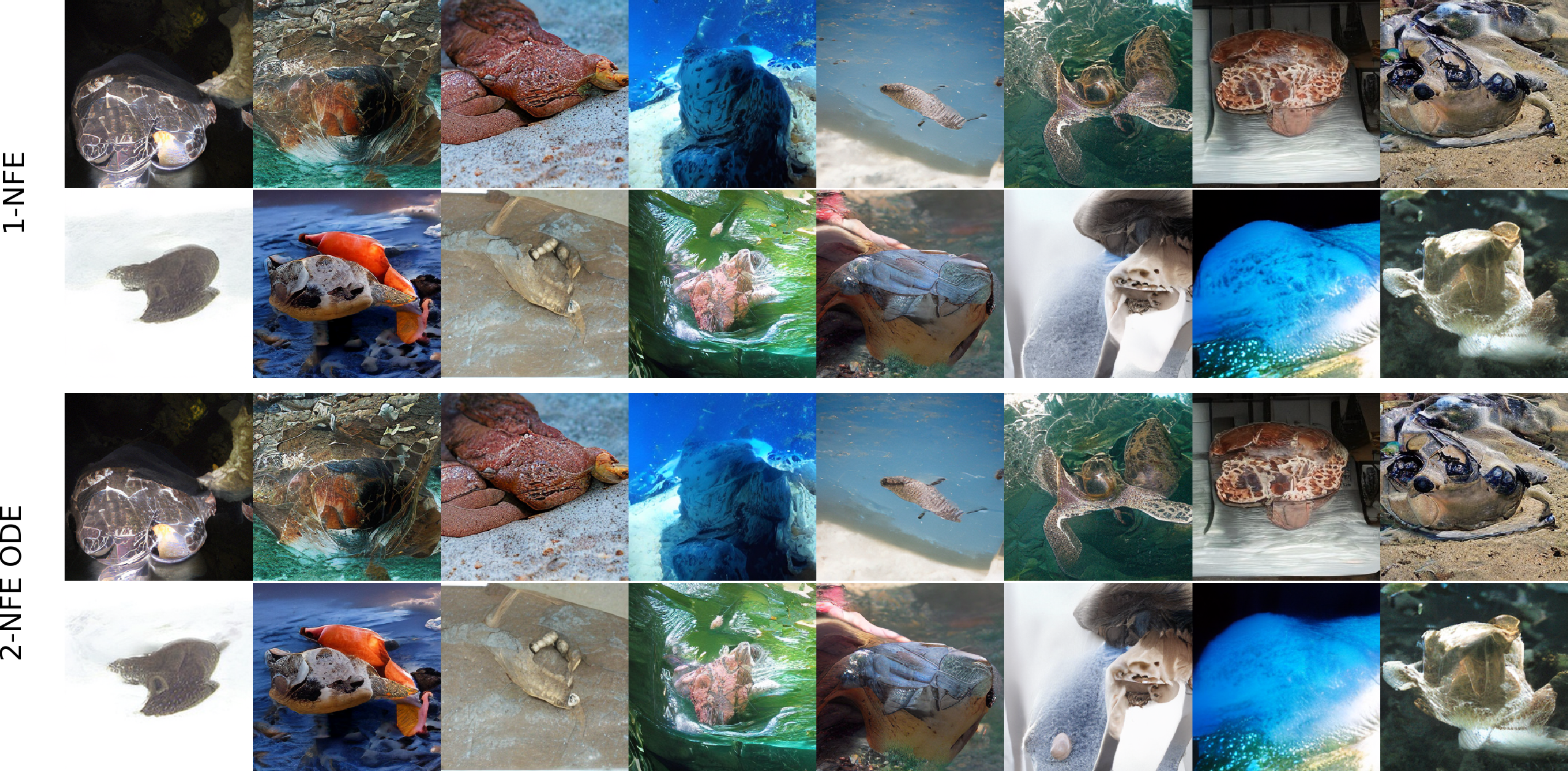}
  \caption{Uncurated EbbFlow samples: ImageNet class 33 (loggerhead),
  seeds $0$--$15$, 400k EMA. Top: one NFE; bottom: two-NFE ODE.}
  \label{fig:samples-loggerhead}
\end{figure}

\begin{figure}[!ht]
  \centering
  \includegraphics[width=0.98\linewidth]{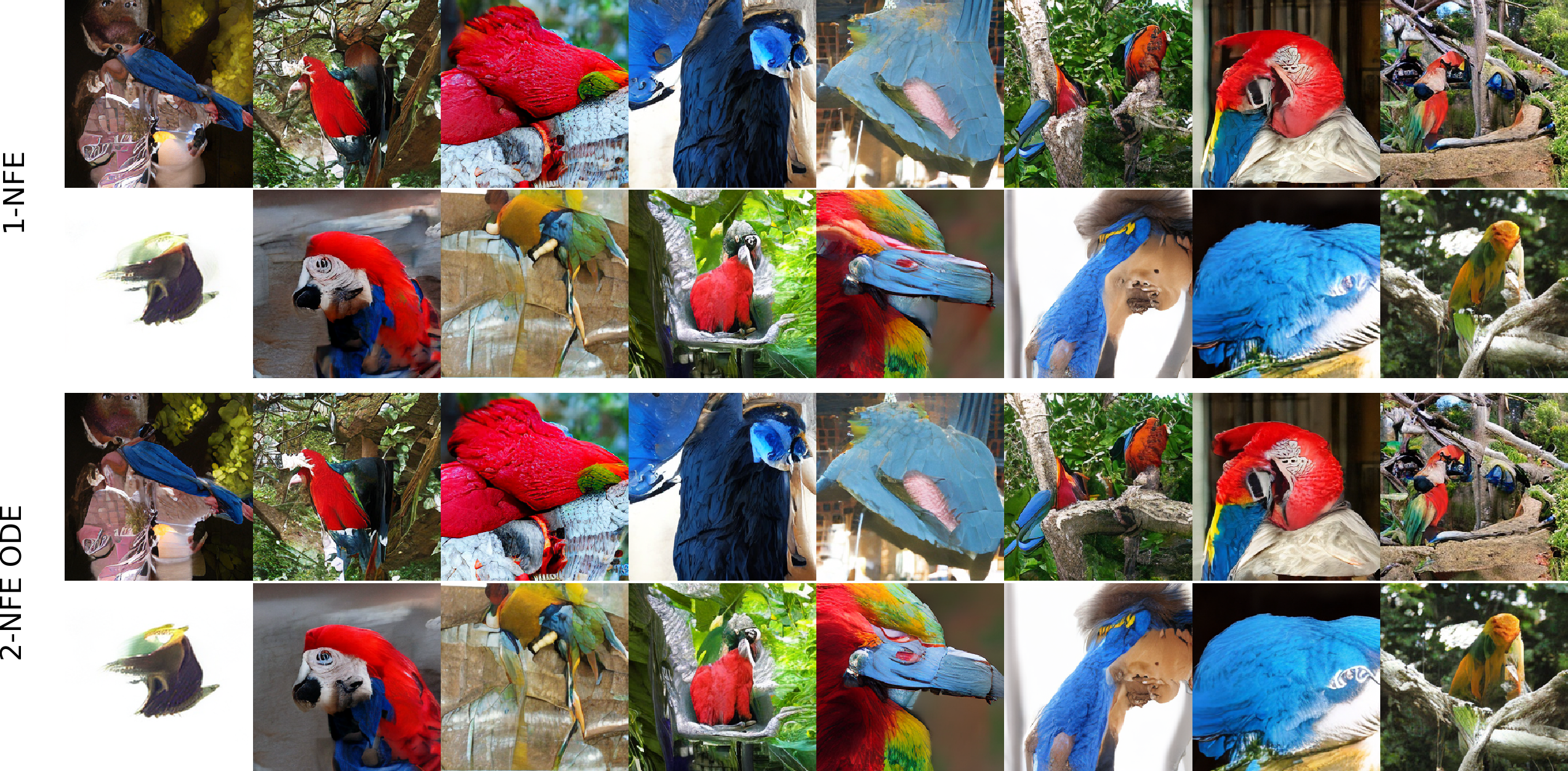}
  \caption{Uncurated EbbFlow samples: ImageNet class 88 (macaw),
  seeds $0$--$15$, 400k EMA. Top: one NFE; bottom: two-NFE ODE.}
  \label{fig:samples-macaw}
\end{figure}

Figures~\ref{fig:samples-loggerhead} and~\ref{fig:samples-macaw} show samples
from the 400k-update EMA checkpoint of the EbbFlow configuration in
Table~\ref{tab:temporal-configurations}, which attains the lowest two-NFE ODE
FID among the reported matched configurations ($36.02$).
For each class, we use all seeds $0$--$15$, without selecting individual outputs,
and keep the class and initial noise fixed between one- and two-NFE generation.
Within each sampler block, seeds appear in row-major order. Both samplers use
guidance scale one; two-NFE ODE uses intermediate time $0.5$.
The full-horizon left-anchor control has the better one-NFE FID; these figures
illustrate EbbFlow's joint curriculum rather than that separate control.

\FloatBarrier
\end{document}